\documentclass[10pt,twocolumn,letterpaper]{article}
\usepackage[pagenumbers]{cvpr}
\ifdefined\pdfsuppressptexinfo
\fi
\usepackage[utf8]{inputenc}
\usepackage[T1]{fontenc}
\usepackage{nicefrac}
\usepackage{microtype}
\usepackage{algorithm}
\usepackage{algorithmic}
\usepackage{multirow}
\usepackage{colortbl}
\usepackage{makecell}
\usepackage{tabularx}

\definecolor{rowblue}{RGB}{236, 245, 255}
\definecolor{tracebg}{RGB}{240, 234, 250}        
\definecolor{traceframe}{RGB}{200, 188, 224}     
\definecolor{promptbg}{RGB}{248, 244, 230}       
\definecolor{promptframe}{RGB}{210, 198, 168}    

\newcommand{\tracebox}[1]{%
  \par\medskip\noindent
  \fcolorbox{traceframe}{tracebg}{%
    \begin{minipage}{\dimexpr\linewidth-2\fboxsep-2\fboxrule\relax}
    \vspace{2pt}\scriptsize
    \setlength{\parindent}{0pt}%
    \setlength{\parskip}{2pt}%
    #1
    \vspace{2pt}
    \end{minipage}%
  }%
  \par\medskip
}

\newcommand{\promptbox}[1]{%
  \par\medskip\noindent
  \fcolorbox{promptframe}{promptbg}{%
    \begin{minipage}{\dimexpr\linewidth-2\fboxsep-2\fboxrule\relax}
    \vspace{2pt}\scriptsize
    \setlength{\parindent}{0pt}%
    \setlength{\parskip}{2pt}%
    #1
    \vspace{2pt}
    \end{minipage}%
  }%
  \par\medskip
}

\newcommand{\videogym}{\textup{\textsc{VideoGym}}}
\newcommand{\videoagent}{\textup{\textsc{VideoAgent}}}
\newcommand{\taskgen}{\textup{\textsc{TaskGenerator}}}
\newcommand{\verifier}{\textup{\textsc{Verifier}}}
\newcommand{\expthree}{\textup{\textsc{EXP3}}}

\definecolor{cvprblue}{rgb}{0.21,0.49,0.74}
\usepackage[pagebackref,breaklinks,colorlinks,allcolors=cvprblue]{hyperref}
\hypersetup{pdftitle={VideoGym: An Adaptive RL Environment for Long-Video Agents}, pdfauthor={Jing Wang, Yi Ke Yun, Yifeng Yao, Huishuai Zhang, Dongyan Zhao, Tao Wang}}
\def\confName{CVPR}
\def\confYear{2027}
\title{\textsc{VideoGym}: An Adaptive RL Environment for Long-Video Agents}
\author{
Jing Wang\textsuperscript{1}\quad
Yi Ke Yun\textsuperscript{2}\quad
Yifeng Yao\textsuperscript{1}\quad
Huishuai Zhang\textsuperscript{1}\quad
Dongyan Zhao\textsuperscript{1}\quad
Tao Wang\textsuperscript{2}\\
\textsuperscript{1}Wangxuan Institute of Computer Technology, Peking University\\
\textsuperscript{2}ByteDance
}

\begin{document}
\maketitle
\begin{abstract}
Reinforcement learning (RL) trains video agents to locate sparse evidence for long-video question answering through structured memory retrieval and temporal re-observation. This training, however, faces a long-tail modality distribution in the data. In our source pool, text-heavy questions account for 58\% of samples, compared with 15\% graph-heavy, 17\% visual-heavy, and 10\% time-range questions. Fixed-distribution RL therefore allocates fewer rollouts to underperforming question types. Reweighting these types could increase their sampling frequency but repeatedly reuses the same finite banks of original questions. To solve this problem, we introduce \videogym{}, a curriculum-based RL environment that jointly constructs video-grounded training tasks and allocates rollouts among them. Within the environment, a memory-augmented \videoagent{} is trained to coordinate retrieval from text, graph, and visual memories with temporal re-observation. Across episodes, \videogym{} adapts \videoagent{}'s training distribution online: it casts per-period sample selection as an adversarial bandit over (modality, mode) arms and uses an EXP3-inspired update to shift sampling toward arms with low answer accuracy.
To supply the augmented-mode arms, a TaskGenerator-Verifier pipeline produces harder variants of existing questions that remain answerable within the same video.
Across five long-video benchmarks, \videogym{} lifts RL accuracy from $60.4\%$ to $64.8\%$ on average and by up to $10.5$\,pp on underrepresented graph-heavy questions with $3.3\times$ sample efficiency.
Together, these results show that task expansion and adaptive allocation can improve video agent learning from limited, imbalanced training data.
\end{abstract}

\section{Introduction}
\label{sec:intro}

\begin{figure*}[t]
    \centering
\captionsetup{skip=6pt}
    \includegraphics[width=\textwidth]{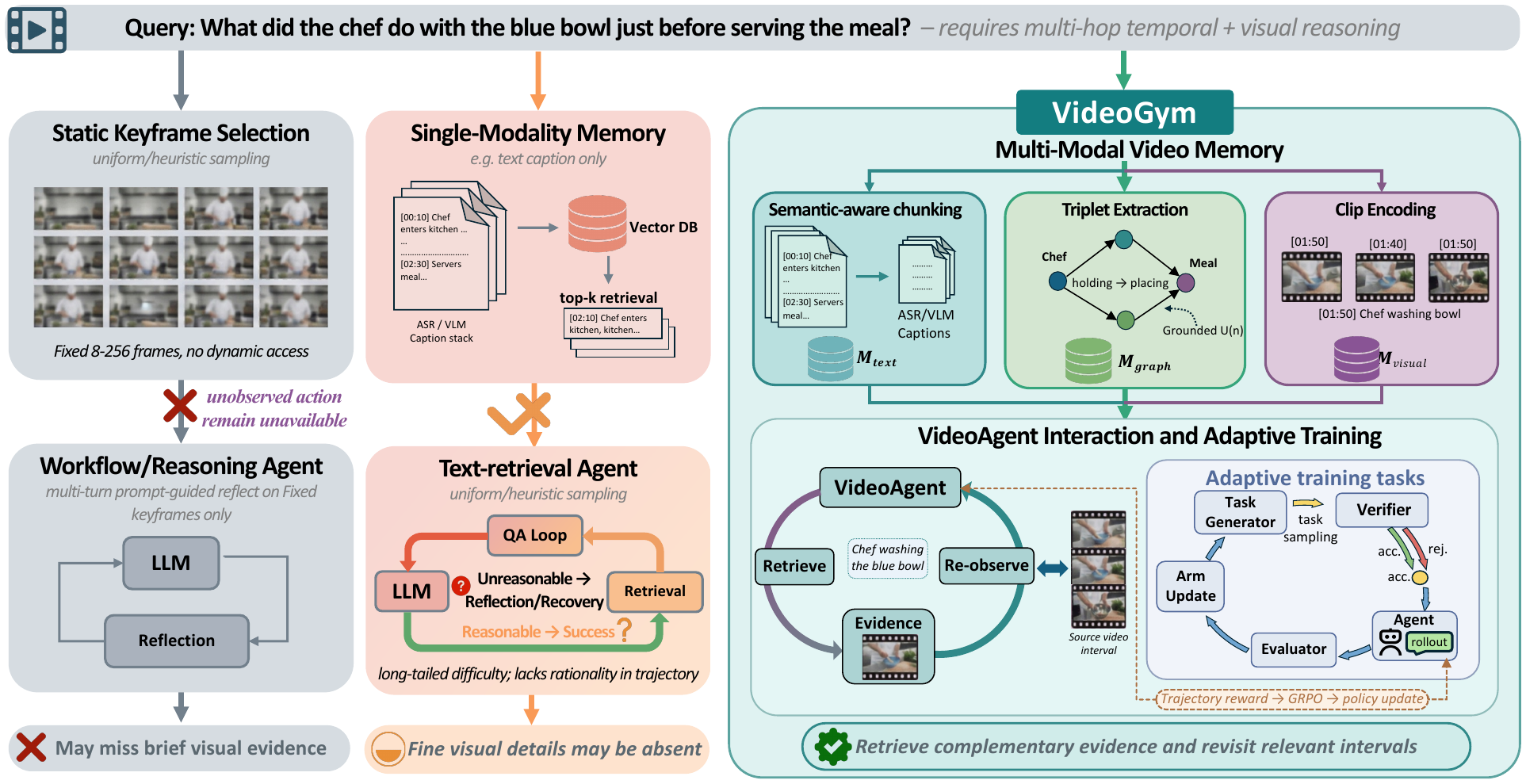}
    \caption{\textbf{Video access and training organization.} Three example access patterns contrast fixed frames with searchable video memory. The right-hand system combines multimodal evidence gathering in \videoagent{} with task construction and feedback-driven allocation in \videogym{}.}
    \label{fig:intro}
\vspace{-6pt}
\end{figure*}

Long-video question answering requires locating sparse evidence across extended recordings. Because a fixed set of initial keyframes may miss brief actions or relations between distant events, memory-augmented approaches organize video content into searchable representations, allowing agents to retrieve relevant evidence on demand and revisit selected video intervals (Figure~\ref{fig:intro})~\citep{long2025seeing,yeo2025worldmm}. Reinforcement learning (RL) trains policies to coordinate these actions over multiple turns, deciding what evidence to acquire next and when to answer~\citep{zhang2026eva,xing2026omniagent}. While these interfaces expand the evidence available to an agent, the questions sampled during RL determine which evidence-acquisition problems the policy learns to solve.

Under fixed-distribution RL, question-pool composition dictates training exposure regardless of agent performance. Our source pool contains 58\% text-heavy questions, compared with 15\% graph-heavy, 17\% visual-heavy, and 10\% time-range questions. The supervised fine-tuned (SFT) policy is already strongest on text-heavy questions and weaker on graph-heavy and visual-heavy groups (Table~\ref{tab:breakdown}). Time-range questions are least frequent yet easier than both groups, so frequency alone does not identify the policy's weaknesses. Reweighting weak groups increases their exposure but reuses the same finite question banks, while a mixture fixed before training cannot track evolving policy performance. Recent work explores adaptive RL environments~\citep{zeng2025rlve,guo2025genenv}, difficulty sampling~\citep{ouyang2025conan}, and failure-driven generation~\citep{zhang2026eva}, yet leaves open a central question for finite video corpora: \textit{How should a video RL environment expand its training tasks and allocate rollouts as the agent's capabilities evolve?}

We introduce \videogym{}, a curriculum-based RL environment that couples video-grounded task construction with adaptive rollout allocation across two interacting levels (Figure~\ref{fig:intro}). Within each episode, its memory-augmented learner, \videoagent{}, queries structured text, graph, and visual memories and uses temporal re-observation to inspect raw frames when retrieved descriptions leave local details unresolved. Across episodes, answer correctness updates task allocation, while composite trajectory rewards train the policy via Group Relative Policy Optimization (GRPO)~\citep{shao2024deepseekmath}.

Task generation expands the available questions, while the curriculum allocates rollouts across question types and original or generated tasks through $(\text{modality}, \text{mode})$ arms. To supply augmented arms, a \taskgen{}--\verifier{} pipeline uses video captions and entity relations to generate variants targeting additional reasoning or finer visual detail, checking each candidate against these representations for answer support and duplication. Across periods, \videogym{} casts arm selection as an adversarial bandit and applies an EXP3-inspired update~\citep{auer2002nonstochastic} to shift rollouts toward low-accuracy arms using outcomes from both task modes.

Across five long-video benchmarks, \videogym{} lifts mean RL accuracy from $60.4\%$ to $64.8\%$ over fixed-distribution RL under matched SFT initialization, action interface, reward, and optimization steps. Gains concentrate on initially weaker question types, reaching $+10.5$\,pp on graph-heavy and $+7.5$\,pp on visual-heavy questions. With generated variants and a fixed rollout budget, \videogym{} matches the 2,000-question baseline using about 600 original RL questions, a $3.3\times$ improvement in original-question efficiency.

Our contributions are threefold:
\begin{itemize}
    \item We develop \videogym{}, a curriculum-based RL environment that trains a memory-augmented \videoagent{} through video-grounded tasks and interactive evidence acquisition.
    \item We couple a source-conditioned \taskgen{}--\verifier{} pipeline with an EXP3-inspired curriculum over $(\text{modality}, \text{mode})$ arms to jointly expand tasks and adapt rollout allocation.
    \item We demonstrate consistent gains across five benchmarks, with the largest improvements on initially weak task groups and $3.3\times$ higher original-question efficiency under a fixed rollout budget.
\end{itemize}

\section{Related Work}
\label{sec:related}
\begin{figure*}[t]
  \centering
\captionsetup{skip=6pt}
  \includegraphics[width=\textwidth]{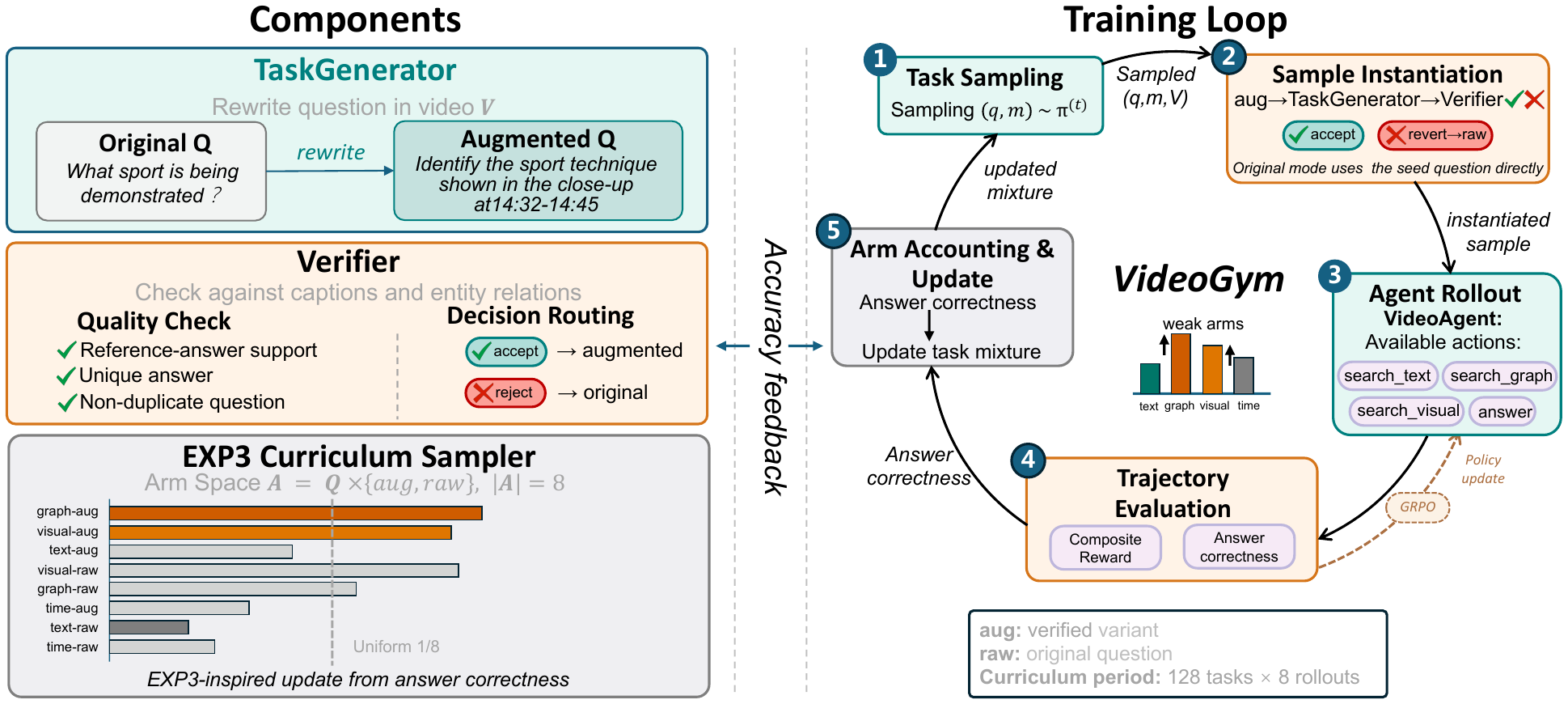}
  \caption{
    \textbf{Task construction and feedback in \videogym{}.} Sampled tasks are instantiated from original questions or verifier-accepted variants. Trajectories provide answer correctness for task allocation and composite rewards for GRPO. Algorithm~\ref{alg:videogym} and Eq.~\eqref{eq:reward} specify these two feedback paths.
  }
  \label{fig:videogym}
\vspace{-6pt}
\end{figure*}

\paragraph{Memory-augmented video agents.}

Recent work has explored structured memory and retrieval mechanisms for long-form video understanding. VideoRAG \citep{ren2025videorag} constructs graph-grounded textual and visual indices for retrieval-augmented generation over hours-long videos. Deep Video Discovery \citep{zhang2025deepvideo} represents videos as a multi-granularity database that can be navigated by an agent loop. M3-Agent \citep{long2025seeing} maintains entity-centric episodic and semantic memory through continuous observation and trains multi-turn memory retrieval with reinforcement learning. WorldMM \citep{yeo2025worldmm} further builds heterogeneous memory across multiple temporal scales for adaptive cross-modal retrieval.

Building on these evidence-access interfaces, \videogym{} studies the training tasks through which an agent learns to use them.

\paragraph{Agentic RL and training environments.}
Research on agent learning spans outcome-based reasoning
\citep{guo2025deepseek}, multi-turn interaction and learning across diverse
environments \citep{yao2022react,xi2025agentgym}, and environment-centric
training where task and feedback generation are co-designed with the policy
\citep{huang2025environment,guo2025genenv,zeng2025rlve,lu2025environment}.
For video agents, Conan~\citep{ouyang2025conan} selects SFT and RL samples by evidence difficulty, combining progressive cold-start training with evidence-based rewards. EVA~\citep{zhang2026eva} expands its GRPO data by using policy failures to guide question generation for additional videos. These approaches connect task design with the acquisition of evidence-gathering policies. \videogym{} couples question expansion within the same videos with online rollout allocation across question types and original or generated tasks. An EXP3-inspired update~\citep{auer2002nonstochastic} uses recent answer correctness to redistribute a fixed rollout budget as the policy evolves.

\section{Method}
\label{sec:method}

\videogym{} is an RL environment for learning to acquire evidence and answer questions about long videos (Figure~\ref{fig:videogym}). In each curriculum period, it samples a batch of tasks, instantiates original or generated questions, and collects groups of evidence-gathering trajectories. Answer correctness updates the task distribution for the next period, while composite trajectory rewards train the policy through GRPO~\citep{shao2024deepseekmath}. We first define the environment and instantiate its learner as \videoagent{}, then describe the curriculum and policy updates.

\subsection{The \videogym{} Environment}
\label{sec:method:formulation}

A training task is denoted by
$s=(\mathcal{V}, x, y, c)$, where $\mathcal{V}$ is a long video,
$x$ is the question, $y$ is the reference answer, and
$c \in \mathcal{C}$ is a question-type label used to organize the task bank. We use
\begin{equation}
\mathcal{C} =
\{\textit{text}, \textit{graph}, \textit{visual}, \textit{time-range}\}.
\end{equation}
We assign these labels from question wording (priority rules in Appendix~\ref{app:impl:data}). Text cues include speech and on-screen text; graph cues include causal, relational, and summarization questions; time-range cues refer to timestamps or event order; and remaining questions enter the visual group. Every group uses the full action interface, allowing the policy to combine multiple evidence sources within one task.

\paragraph{Episode-level interaction.}
The policy $\pi_\theta$ receives a question $x$ and an initial keyframe panel, and can query the video's memory $\mathcal{M}(\mathcal{V})$ or request new frames from a selected interval through actions $u_k\in\mathcal{A}_{\mathrm{agent}}$. The video remains fixed, while each action changes the evidence available to the policy. At turn $k$, conditioned on the initial context and interaction history
$h_{<k} = (u_1, e_1, \ldots, u_{k-1}, e_{k-1})$, the policy selects $u_k$ and
either receives evidence $e_k$ or terminates the episode by emitting
$u_T = \texttt{answer}$ together with the prediction $\hat{y}$. The
resulting evidence trajectory has length $T \le T_{\max}$:
\begin{equation}
  \tau = (u_1, e_1, \ldots, u_{T-1}, e_{T-1}, u_T, \hat{y}).
\end{equation}
An answer ends the episode, whereas reaching the turn budget $T_{\max}$ triggers an answer from the current evidence. Comparing the final prediction $\hat{y}$ with the reference answer $y$ yields binary correctness, which guides task allocation and contributes to the composite trajectory reward (Section~\ref{sec:method:training}).

\paragraph{Curriculum-level decision.}
At each curriculum period, the environment selects the task instances used to train $\pi_\theta$. We define the curriculum arm space
\begin{equation}
  \mathcal{A}_{\mathrm{gym}} = \mathcal{C} \times \{\textit{raw}, \textit{aug}\},
\end{equation}
with arms $z = (c, m) \in \mathcal{A}_{\mathrm{gym}}$. The first factor selects a question type $c$. The second selects an existing instance from $\mathcal{B}_c$ ($m = \textit{raw}$) or a generation--verification procedure that falls back to the seed if verification fails ($m = \textit{aug}$). In each period, the curriculum samples arms from $p^{(t)}$ to instantiate tasks for the agent (Section~\ref{sec:method:videogym}).

\subsection{\videoagent{}}
\label{sec:method:videoagent}

\videoagent{} instantiates the policy and video-access interface in \videogym{} (Figure~\ref{fig:intro}). Its planner chooses among complementary memory queries and temporal re-observation, then integrates the returned evidence across turns.

\paragraph{Video memory.}
For each video $\mathcal{V}$, \videoagent{} builds a cached external
memory
\begin{equation}
\mathcal{M}(\mathcal{V}) =
(\mathcal{M}_{\mathrm{text}},
 \mathcal{M}_{\mathrm{graph}},
 \mathcal{M}_{\mathrm{visual}}),
\end{equation}
which is constructed once offline and reused during SFT, RL rollouts,
and evaluation.
The three stores are organized around the kinds of evidence that
long-video questions require.
$\mathcal{M}_{\mathrm{text}}$ indexes segment-level captions and ASR
transcripts, covering evidence that is explicitly described in speech,
narration, or generated captions.
$\mathcal{M}_{\mathrm{graph}}$ lifts caption-derived entities and
relations into a linked graph, allowing the planner to track recurring
entities and reason over relations that appear in non-contiguous
segments.
$\mathcal{M}_{\mathrm{visual}}$ stores timestamped segment-level visual
embeddings for locating appearance-level evidence. These stores give the planner complementary entry points into a long recording, while temporal re-observation lets it inspect a selected interval when retrieved descriptions or visual anchors leave a detail unresolved.

\paragraph{Action interface.}
The planner action space is
\begin{equation}
\begin{aligned}
\mathcal{A}_{\mathrm{agent}} = \{&\texttt{search\_text},\ \texttt{search\_graph},\\
&\texttt{search\_visual},\\
&\texttt{search\_time\_range},\ \texttt{answer}\}.
\end{aligned}
\end{equation}
The first three actions query the corresponding memory store for textual, relational, or visual evidence. The action \texttt{search\_time\_range}
takes a temporal window and returns newly sampled frames from that interval, exposing local visual detail through direct access to the source video.
The action \texttt{answer} terminates the episode and emits $\hat{y}$.

\paragraph{Planner loop.}
At each turn, \videoagent{} produces a planner state and one structured action tag, which the environment executes against the fixed memory indices or source video. We apply the same rollout constraints during SFT trace collection, RL training, and evaluation: repeated retrieval queries are blocked to avoid loops, malformed action tags trigger a retry or a blocked-action record, and the episode is bounded by the turn budget $T_{\max}$.

\subsection{Adaptive Task Curriculum}
\label{sec:method:videogym}

Allocating more training to an underperforming question type increases reuse of its finite original-question bank. The raw and augmented arms let the curriculum allocate rollouts to existing questions or generate additional tasks from the same video representations, updating the mixture for the next period from answer correctness.

\paragraph{Source-conditioned task generation.}
For an augmented task, \taskgen{} receives a seed $s=(\mathcal{V},x,y,c) \in \mathcal{B}_c$, the video's captions, and its entity relations, and proposes $\tilde{s}=(\mathcal{V},\tilde{x},\tilde{y},c)$, preserving the question type while seeking an additional reasoning step or finer detail. \verifier{} screens the candidate for well-formedness, answer support in the supplied representations, and duplication (prompt schemas in Appendices~\ref{app:prompts:gen} and~\ref{app:prompts:verify}).

If verification rejects a candidate, the environment executes its original seed question. Curriculum feedback remains associated with the sampled proposal arm, since the augmented arm selects the full construction procedure, including fallback. GRPO uses the instantiated question and its reference answer; accepted-augmentation exposure counts only tasks that execute accepted variants. An unsupported reference can make a task appear difficult despite useful retrieval (Appendix~\ref{app:cases:fail}).

\begin{table*}[t]
\centering
\caption{\textbf{Main results on five long-video benchmarks.} Accuracy (\%). Baselines are re-evaluated from released code and checkpoints under a common multiple-choice protocol. Bold indicates the best result per column; \colorbox{rowblue}{blue rows} show our training stages. For RL, Avg. reports the mean and sample standard deviation of the five-benchmark average across three seeds.}
\label{tab:main}
\begingroup
\small
\setlength{\tabcolsep}{3.6pt}
\begin{tabular}{@{}l r rrrrr r@{}}
\toprule
\textbf{Method} & \textbf{Size} & \textbf{\makecell{EgoLife\\QA}} & \textbf{\makecell{Ego-R1\\Bench}} & \textbf{\makecell{Hippo\\Vlog}} & \textbf{\makecell{LV\\Bench}} & \textbf{\makecell{Video-MME\\(Long)}} & \textbf{Avg.}\\
\midrule
\multicolumn{8}{l}{Long-video models}\\
VideoChat-Flash~\citep{li2025videochatflash} & 7B & 34.6 & 41.3 & 61.2 & 47.9 & 54.8 & 48.0 \\
Time-R1~\citep{wang2025timer1} & 7B & 45.1 & 43.6 & 57.4 & 40.2 & 52.7 & 47.8 \\
Video-RTS~\citep{wang2025videorts} & 7B & 45.7 & 44.8 & 62.3 & 45.1 & 54.4 & 50.5 \\
\midrule
\multicolumn{8}{l}{RAG-based methods}\\
LightRAG~\citep{guo2024lightrag} & -- & 43.2 & 42.5 & 47.1 & 31.6 & 42.4 & 41.4 \\
Video-RAG~\citep{luo2024video} & 7B & 43.9 & 41.7 & 59.3 & 38.5 & 54.9 & 47.7 \\
HippoRAG~\citep{gutierrez2024hipporag} & -- & 55.2 & 52.4 & 60.6 & 45.3 & 48.7 & 52.4 \\
\midrule
\multicolumn{8}{l}{Memory and retrieval agents}\\
EgoRAG~\citep{yang2025egolife} & -- & 36.1 & 40.4 & 56.2 & 34.6 & 43.5 & 42.2 \\
Conan~\citep{ouyang2025conan} & 7B & 38.7 & 40.9 & 63.4 & 38.8 & 52.6 & 46.9 \\
M3-Agent~\citep{long2025seeing} & 8B & 50.4 & 47.6 & 66.2 & 43.7 & 50.1 & 51.6 \\
Ego-R1~\citep{tian2025egor1} & 3B & 44.6 & 45.9 & 64.3 & 43.5 & 63.8 & 52.4 \\
HippoMM~\citep{lin2025hippomm} & -- & 52.3 & 50.6 & \textbf{77.2} & 42.4 & 48.3 & 54.2 \\
WorldMM-8B~\citep{yeo2025worldmm} & 8B & 53.6 & 49.5 & 67.4 & 52.3 & 63.5 & 57.3 \\
\midrule
\rowcolor{rowblue}
\multicolumn{8}{l}{Ours: training progression}\\
\rowcolor{rowblue}
\videoagent{} (training-free) & 8B & 54.1 & 51.5 & 66.8 & 49.1 & 54.0 & 55.1 \\
\rowcolor{rowblue}
+ SFT & 8B & 55.8 & 54.3 & 68.2 & 51.8 & 59.4 & 57.9 \\
\rowcolor{rowblue}
+ RL (static tasks) & 8B & 57.4 & 55.6 & 71.5 & 53.7 & 64.0 & $60.4\pm0.5$ \\
\rowcolor{rowblue}
+ RL (\videogym{}) & 8B & \textbf{61.5} & \textbf{59.9} & 75.5 & \textbf{58.3} & \textbf{69.0} & $\mathbf{64.8}\pm0.6$ \\
\bottomrule
\end{tabular}
\endgroup
\end{table*}

\begin{table*}[t]
\centering
\caption{
  \textbf{Accuracy by question type.} RL rows are three-seed means.
  $\Delta$ gives the gain of RL (\videogym{}) over RL (static tasks).
  Bold gains exceed 5 percentage points. Question types follow the task-bank labeling protocol.
}
\label{tab:breakdown}
\begingroup
\small
\begin{tabular}{lcccc}
\toprule
\textbf{Method}
  & \textbf{Text-heavy} & \textbf{Graph-heavy}
  & \textbf{Visual-heavy} & \textbf{Time-range} \\
\midrule
Training-free          & 67.3 & 41.8 & 43.9 & 53.2 \\
$+$ SFT                & 70.1 & 44.2 & 46.8 & 56.4 \\
$+$ RL (static tasks) & 72.6 & 47.0 & 49.5 & 58.8 \\
$+$ RL (\videogym{})  & 73.9 & 57.5 & 57.0 & 64.2 \\
\midrule
$\Delta$ & $+$1.3 & $\mathbf{+10.5}$ & $\mathbf{+7.5}$ & $\mathbf{+5.4}$\\
\bottomrule
\end{tabular}
\endgroup
\end{table*}

\paragraph{Performance-driven sampling.}
During each curriculum period $t$, \videogym{} draws $B=128$ proposal arms $a_i\sim p^{(t)}$, instantiates one task per draw, and collects $G=8$ trajectories per task with $p^{(t)}$ fixed. The recorded $a_i$ is unchanged by fallback. Correctness is $\rho_{ij}=\mathbf{1}[\textsc{Match}(\hat{y}_{ij},y_i^{\star})=1]$, where $y_i^{\star}$ is the instantiated task's reference answer. The $BG=1024$ outcomes update the next period's distribution via an EXP3-inspired rule~\citep{auer2002nonstochastic}.

\videogym{} maintains weights $\{w_z^{(t)}\}$ and distribution
$p^{(t)}$ over $\mathcal{A}_{\mathrm{gym}}$, initialized uniformly with
$w_z^{(0)} = 1$ and $p_z^{(0)} = 1/K$, where
$K = |\mathcal{A}_{\mathrm{gym}}| = |\mathcal{C}| \times 2 = 8$. Let $N_z^{(t)}=G\sum_{i=1}^{B}\mathbf{1}[a_i=z]$ count trajectories from proposal arm $z$. For $N_z^{(t)}>0$, its mean correctness is
$\bar{\rho}_z^{(t)} = (N_z^{(t)})^{-1} \sum_{i: a_i = z}\sum_{j=1}^{G} \rho_{ij}$. We cap the arm's mean error, then aggregate it using its batch frequency and proposal probability:
\begin{gather}
  \begin{aligned}
  \ell_z^{(t)} &= \min\!\bigl(1 - \bar{\rho}_z^{(t)},\, \delta\bigr),\\
  \tilde{\ell}_z^{(t)} &= \frac{N_z^{(t)}}{BG\,p_z^{(t)}}\ell_z^{(t)},
  \end{aligned}\label{eq:weakness}\\
  w_z^{(t+1)} = w_z^{(t)} \exp\!\bigl(\gamma\, \tilde{\ell}_z^{(t)} / K\bigr), \label{eq:weight_update}\\
  p_z^{(t+1)} = (1-\gamma)\,
    \frac{w_z^{(t+1)}}{\sum_{z'} w_{z'}^{(t+1)}} + \frac{\gamma}{K}, \label{eq:prob_update}
\end{gather}
with $\gamma = 0.1$ and $\delta = 0.80$. For unobserved arms, $\tilde{\ell}_z^{(t)}=0$ and weights remain unchanged. The factor $N_z^{(t)}/(BG)$ accounts for observed batch frequency. The cap and exploration floor $p_z^{(t)}\ge\gamma/K$ bound each period's log-weight increment by $\delta$ (Appendix~\ref{app:algo:cap}). Algorithm~\ref{alg:videogym} summarizes the loop; Sections~\ref{sec:exp:longtail} and~\ref{sec:exp:ablation} evaluate allocation and augmentation.

\subsection{Policy Optimization}
\label{sec:method:training}

\paragraph{Two-stage SFT initialization.}
We initialize the policy through two stages of supervised fine-tuning. Planner SFT supervises the next structured action from a question and partial evidence history, while Answer SFT supervises the final response from the accumulated evidence.

\paragraph{GRPO with composite trajectory rewards.}
Let $\mathcal{A}_{\mathrm{ret}} = \mathcal{A}_{\mathrm{agent}} \setminus \{\texttt{answer}\}$ contain the four retrieval actions, including temporal re-observation. The number of retrieval turns is
\begin{equation}
T_{\mathrm{search}}
=
\sum_{k=1}^{T-1}
\mathbf{1}[u_k \in \mathcal{A}_{\mathrm{ret}}].
\end{equation}
GRPO updates the policy from the group of $G$ trajectories sampled for each task using the composite reward
\begin{equation}
\begin{aligned}
R(\tau) ={}& \rho\bigl(1 + \lambda_e\, b_{\mathrm{evid}}\\
 &\quad - \lambda_{\mathrm{lazy}}\,\mathbf{1}[T_{\mathrm{search}}=0]\bigr)\\
 &- \lambda_{\mathrm{eff}}\, c_{\mathrm{eff}},
\end{aligned}
\label{eq:reward}
\end{equation}
where $b_{\mathrm{evid}}\in[0,1]$ is the fraction of retrieval turns whose returned evidence matches the content of the correct answer option under rule-based matching, gated by correctness $\rho$. The indicator $\mathbf{1}[T_{\mathrm{search}}=0]$ penalizes correct answers produced without retrieval, and $c_{\mathrm{eff}}\in[0,1]$ combines normalized turn count, forced termination, blocked actions, and format retries. We use $\lambda_e=\lambda_{\mathrm{eff}}=0.3$ and $\lambda_{\mathrm{lazy}}=0.2$ (cost decomposition in Appendix~\ref{app:algo:reward}).

The evidence and cost terms can distinguish trajectories with identical answer correctness when their evidence use or interaction cost differs, providing additional within-group reward variation for GRPO.

\section{Experiments}
\label{sec:experiments}

We evaluate overall accuracy across five benchmarks, and then examine task allocation, task expansion and validity, and evidence acquisition.

\begin{figure*}[t]
\centering
\captionsetup{skip=6pt}
\includegraphics[width=0.85\textwidth]{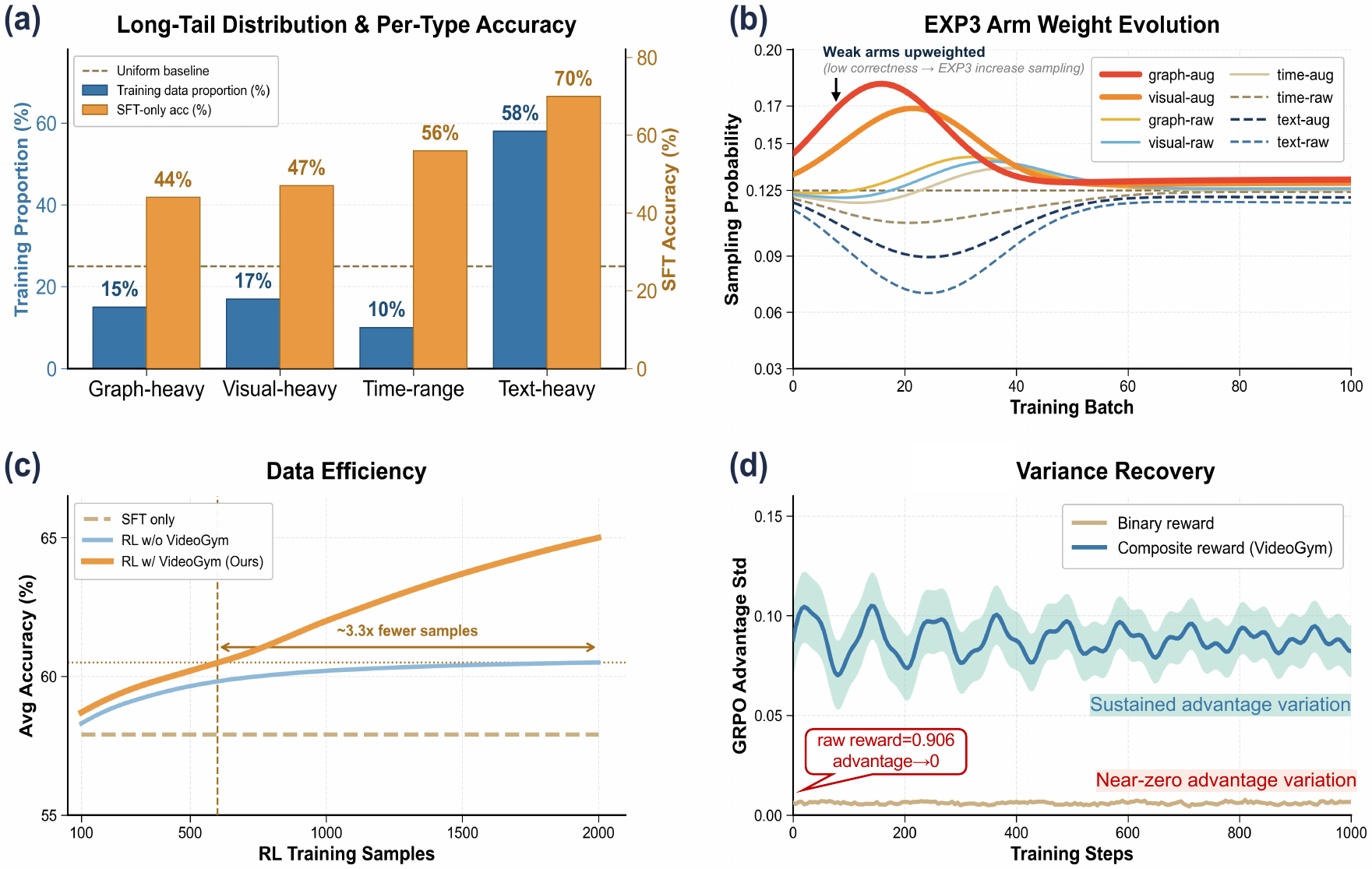}
\caption{
  \textbf{Task allocation and policy learning in \videogym{}.}
  \textbf{(a)} Source-pool distribution and SFT accuracy by question type.
  \textbf{(b)} Curriculum arm probabilities from one training run.
  \textbf{(c)} Accuracy versus the number of unique raw questions at a fixed rollout budget, with generated variants included in \videogym{} training.
  \textbf{(d)} GRPO advantage standard deviation under binary and composite rewards. Both main RL settings use the composite reward; Appendix~\ref{app:exp:reward} discusses this shared optimization signal.
}
\label{fig:analysis}
\vspace{-6pt}
\end{figure*}

\subsection{Experimental Setup}
\label{sec:exp:setup}

\paragraph{Benchmarks and baselines.}
We evaluate on five long-video benchmarks, covering broad video understanding with LVBench~\citep{wang2024lvbench} and Video-MME (Long)~\citep{fu2024video}, and egocentric, memory-intensive, and multi-hop reasoning with EgoLifeQA~\citep{yang2025egolife}, Ego-R1 Bench~\citep{tian2025egor1}, and HippoVlog~\citep{lin2025hippomm}.
We compare against long-video LLMs, RAG-based methods, and memory-based video agents, re-evaluating all baselines from released code and checkpoints under a common multiple-choice protocol.
Question-type results use the same question-side labeling protocol as the training banks $\{\mathcal{B}_c\}_{c\in\mathcal{C}}$, with the full action interface shared across all groups.

\paragraph{Training data.}
SFT and RL use the multiple-choice portion of LLaVA-Video-178K~\citep{zhang2024video} (Appendix~\ref{app:impl:data}).
In the source question pool, text-heavy questions account for $58\%$, compared with $17\%$ visual-heavy, $15\%$ graph-heavy, and $10\%$ time-range questions.
A GPT-4o-backed teacher produces multi-turn trajectories, of which approximately 96K pass correctness and structural filtering for Planner SFT followed by Answer SFT.
RL uses approximately 2K seed prompts sampled from the accepted pool.
Fixed-distribution RL samples these prompts at their natural distribution, whereas \videogym{} uses the same seed questions, generates variants, and adjusts sampling over raw and augmented arms using answer feedback.

\paragraph{Controlled RL comparison.}
All RL variants start from the same SFT checkpoint, retain the same memory and temporal re-observation interface, and use GRPO~\citep{shao2024deepseekmath} with the composite reward in Eq.~\eqref{eq:reward}.
The two main RL settings also match optimization steps and evaluate each run's final checkpoint.

\paragraph{Implementation.}
The agent is initialized from Qwen3-VL-8B~\citep{bai2025qwen3}.
Video memory uses Whisper~\citep{radford2023robust} for ASR, Qwen3-VL for captions, and ImageBind~\citep{girdhar2023imagebind} for visual embeddings.
Each SFT stage runs for 2 epochs, and RL uses group size $G=8$, turn limit $T_{\max}=10$, reward weights $\lambda_e=\lambda_{\mathrm{eff}}=0.3$, and curriculum exploration rate $\gamma=0.1$ on one $8{\times}$H100-80G node (Appendix~\ref{app:algo:hparams}).

\subsection{Long-Video Question Answering}
\label{sec:exp:main}

Under matched training conditions, \videogym{} raises mean accuracy from $60.4\%$ to $64.8\%$ across the five benchmarks (Table~\ref{tab:main}), with gains over static-task RL ranging from $4.0$ percentage points (pp) on HippoVlog to $5.0$\,pp on Video-MME (Long). Given the shared initialization, interface, reward, and optimization budget in Section~\ref{sec:exp:setup}, this improvement reflects the effect of changing task construction and rollout allocation. Building on the training-free ($55.1\%$) and SFT ($57.9\%$) stages, the resulting system achieves the highest five-benchmark average among the compared methods, $7.5$\,pp above WorldMM-8B.

The gains span both egocentric benchmarks and broader long-video evaluations. Their relatively narrow range across datasets complements the larger variation across question types in Table~\ref{tab:breakdown}, motivating a closer examination of how the training benefits are distributed across the agent's tasks.

\subsection{Task Allocation during Learning}
\label{sec:exp:longtail}

Fixed-distribution RL improves all four question groups by similar amounts over SFT, preserving their initial performance gaps (Table~\ref{tab:breakdown}). \videogym{} yields its largest gains on the initially weaker graph-heavy and visual-heavy groups, at $+10.5$ and $+7.5$\,pp, while also improving time-range ($+5.4$\,pp) and text-heavy ($+1.3$\,pp) accuracy to narrow the gaps across groups.

Source frequency and initial accuracy rank the question types differently: time-range questions are least frequent but exceed graph-heavy and visual-heavy questions in SFT accuracy (Figure~\ref{fig:analysis}(a)). During training, the curriculum initially emphasizes the \texttt{graph-aug} and \texttt{visual-aug} arms before moving back toward uniform allocation (Figure~\ref{fig:analysis}(b)), and these are also the groups with the largest final gains. Replacing this adaptive schedule with uniform sampling over the raw and augmented arms reaches $62.2\%$, compared with $64.8\%$ for the full system (Table~\ref{tab:gym_ablation}).

\begin{table}[t]
\centering\small
\caption{\textbf{Allocation on a shared task bank.} Five-benchmark accuracy across three training seeds, relative training tokens, and accepted-augmentation exposure. All rows match the rollout budget. The pilot-average diagnostic uses one additional pilot run.}
\label{tab:allocation_control}
\setlength{\tabcolsep}{3pt}
\begin{tabular}{@{}lccc@{}}\toprule
Allocation & Accuracy & Tokens & Aug. (\%)\\\midrule
Uniform & $62.02\pm0.44$ & 0.987 & 41.6\\
Fixed initial-error & $63.33\pm0.40$ & 1.000 & 45.8\\
Fixed pilot-average & $63.85\pm0.42$ & 1.014 & 52.7\\
\videogym{} & $64.37\pm0.34$ & 1.021 & 53.2\\
\bottomrule\end{tabular}
\end{table}

\paragraph{Feedback beyond a fixed mixture.}
To isolate rollout allocation, we train all methods on the same preconstructed bank of original questions and generated proposals (Table~\ref{tab:allocation_control}). Each method collects fresh trajectories under matched initialization, GRPO settings, and rollout counts (Appendix~\ref{app:allocation-control}). Adaptive allocation exceeds a mixture fixed from initial SFT errors by $1.0$\,pp, and also exceeds a mixture fixed from an independent adaptive pilot's average proposal distribution by $0.5$\,pp.

The pilot-average baseline already incorporates the mean allocation preferences of an adaptive training run. Its accepted-augmentation exposure and token consumption are close to those of \videogym{}. The remaining advantage is consistent with the value of adapting task allocation over training, beyond selecting a favorable overall mixture.

\begin{table*}[t]
\begin{minipage}[t]{0.60\textwidth}
\centering
\caption{
  \textbf{\videoagent{} inference-time ablations.}
  Memory-ablation rows disable one store; ``single-shot'' removes the planner loop.
  All rows use one model trained with \videogym{}; the main-table RL averages instead report three-seed means.
  Avg. is over the two benchmarks shown.
}
\label{tab:va_ablation}
\footnotesize
\setlength{\tabcolsep}{2.5pt}
\begin{tabular}{lcc c}
\toprule
\textbf{Configuration} & \textbf{EgoLifeQA} & \textbf{LVBench} & \textbf{Avg.}\\
\midrule
Full system (RL (\videogym{})) & 62.1 & 58.7 & 60.4\\
\midrule
w/o Graph Memory                  & 60.3 & 55.2 & 57.8\\
w/o Visual Memory                 & 61.2 & 56.8 & 59.0\\
w/o Text Memory                   & 58.4 & 51.2 & 54.8\\
Single-shot (no planner)          & 58.9 & 52.4 & 55.7\\
\bottomrule
\end{tabular}
\end{minipage}\hfill
\begin{minipage}[t]{0.37\textwidth}
\centering
\caption{
  \textbf{\videogym{} component ablations.} Five-benchmark average accuracy.
  The three component ablations each use one run; fixed-distribution RL and the full system are three-seed means.
  Uniform sampling replaces adaptive allocation; the other ablations remove augmentation or verification.
}
\label{tab:gym_ablation}
\footnotesize
\setlength{\tabcolsep}{3pt}
\begin{tabular}{lc}
\toprule
\textbf{Configuration} & \textbf{Avg.}\\
\midrule
RL (static tasks) & 60.4\\
\videogym{} w/ uniform sampling & 62.2\\
\videogym{} w/o augmentation & 63.4\\
\videogym{} w/o verification & 64.3\\
\videogym{} full (ours) & \textbf{64.8}\\
\bottomrule
\end{tabular}
\end{minipage}
\end{table*}

\begin{table}[t]
\centering\small
\caption{\textbf{Paired evidence acquisition.} The same 160 held-out questions, initial observations, and observation limits. Acc. is answer accuracy; Suff. is source-verified evidence sufficiency; Joint requires both.}
\label{tab:paired_evidence}
\begin{tabular}{@{}lccc@{}}\toprule
Training & Acc. (\%) & Suff. (\%) & Joint (\%)\\\midrule
Fixed initial-error & 61.9 & 75.6 & 59.4\\
\videogym{} & 69.4 & 83.1 & 67.5\\
\bottomrule\end{tabular}
\end{table}

\subsection{Task Expansion and Source-Video Validity}
\label{sec:exp:ablation}

Without augmentation, adaptive allocation reaches $63.4\%$, compared with $60.4\%$ for fixed-distribution RL (Table~\ref{tab:gym_ablation}). Adding generated variants raises accuracy to $64.8\%$, whereas removing verification yields $64.3\%$. Under the same SFT initialization and rollout budget, \videogym{} also matches the accuracy of the 2,000-question fixed-distribution baseline using about 600 original questions together with generated variants (Figure~\ref{fig:analysis}(c)), a $3.3\times$ improvement in original-question efficiency at a fixed rollout budget.

\paragraph{What the generated questions add.}
Two annotators independently audited 200 accepted variants and their seeds against source videos (50 pairs per type, one pair per video), identifying the video-supported answer before seeing the proposed reference (Appendix~\ref{app:source-audit}). The audit finds 194 of the 200 accepted variants and 199 of the 200 seed questions valid. Among the 194 valid variants, 118 require locating additional moments, relations, or visual details beyond the seed question, whereas 43 change the task without expanding the required evidence and 33 reuse the seed's answer and evidence.

The distinction concerns the evidence the policy must acquire. A new question can reuse the seed's observations or require the agent to locate additional information in the same recording. The latter expands the acquisition problems available for training from a fixed video corpus. We next examine whether the trained policy more often acquires sufficient evidence under matched observation limits.

\subsection{Evidence Acquisition}
\label{sec:exp:curriculum}

\paragraph{Evidence interface.}
Inference-time ablations on EgoLifeQA and LVBench test which evidence resources the trained policy uses (Table~\ref{tab:va_ablation}). Removing text, graph, or visual memory lowers average accuracy by $5.6$, $2.6$, and $1.4$\,pp, respectively, while removing the planner loop lowers it by $4.7$\,pp. The case studies in Appendix~\ref{app:cases} show how these resources interact across turns: an unsuccessful text query leads to visual retrieval for an on-screen text question, while an object-function question combines text, graph, and visual evidence. Thus, a question-type label can lead to multiple evidence paths.

\paragraph{Behavior under matched observation limits.}
We evaluate one fixed initial-error and adaptive checkpoint pair from the shared-bank control on 160 held-out questions with identical observation limits. Annotators verify tool returns against source videos, blinded to policy identities (Table~\ref{tab:paired_evidence}; Appendix~\ref{app:paired-evidence}). Correct answers accompanied by sufficient evidence increase by $8.1$\,pp overall (95\% CI $[0.6,15.6]$). Mean frame use ($46.8$ vs.\ $48.1$) and input-token use ($10.94$K vs.\ $11.21$K) remain comparable across the two policies.

These evaluations address complementary aspects of evidence acquisition.
The inference-time ablations measure the policy's reliance on memory stores and planning, whereas the checkpoint comparison examines how training changes evidence acquisition through the same interface.
Answer accuracy alone does not establish whether the available observations support the prediction.
Joint success additionally requires source-verified evidence sufficiency.
This distinction is particularly informative on the 96 questions whose initial observations are insufficient (Appendix~\ref{app:paired-evidence}).
On this subset, \videogym{} raises evidence sufficiency from $59.4\%$ to $71.9\%$ and joint success from $49.0\%$ to $59.4\%$.
These gains show that the improvement extends to obtaining supporting evidence during interaction, connecting the training comparison to the agent's ability to resolve gaps in its initial observations.

\section{Conclusion}
\label{sec:conclusion}

\videogym{} integrates source-conditioned task construction,
multi-turn video interaction, and feedback-driven rollout allocation
within a single RL environment. With the agent interface and reward
held fixed, it raises five-benchmark average accuracy from $60.4\%$
to $64.8\%$, with larger gains on underrepresented question types
and reduced reliance on original RL questions.
Shared-bank allocation comparisons further favor feedback-driven sampling
over both initial-error and pilot-informed fixed mixtures.
Source-video auditing supports the validity of accepted tasks and
quantifies near-duplication, while paired evaluation shows more correct answers
accompanied by sufficient evidence. These findings highlight the
training task distribution as a design choice for video-agent RL,
with task construction and allocation shaping the evidence-acquisition
problems encountered during learning.

\paragraph{Limitations.}
Task generation relies on offline captions and entity relations,
which constrain the video information available for constructing
new questions. The predefined question-type partition may also
overlook differences in evidence requirements within each type.
Our source-video audit covers a limited sample of questions, and the
paired behavioral evaluation uses one checkpoint pair. Evaluations
across additional banks and training runs are needed to assess
the consistency of these diagnostic findings.
Appendix~\ref{app:disc} discusses further limitations and broader impacts.

\label{page:main-body-end}
\clearpage
\typeout{VIDEOGYM_MAIN_BODY_PAGES=\the\numexpr\value{page}-1\relax}
\label{page:llm-statement-start}
\section*{LLM Usage Statement}
\label{sec:ai-use}

Generative models are used in our research pipeline to produce video captions
for memory construction (Qwen3-VL), generate teacher trajectories for supervised
fine-tuning (GPT-4o), and generate and filter augmented training questions through
\taskgen{} and \verifier{} (GPT-4o). These uses are described in the method and
implementation sections and in the prompt schemas
(Appendices~\ref{app:impl} and~\ref{app:prompts}).
Generative AI tools were also used to assist with paper writing and editing.
The authors are responsible for the final text, methodology, results and references.

\label{page:references-start}
{
  \small
  \bibliographystyle{ieeenat_fullname}
  \bibliography{main}
}
\clearpage
\typeout{VIDEOGYM_SUBMISSION_PAGES=\the\numexpr\value{page}-1\relax}
\onecolumn
\setcounter{page}{1}
\renewcommand{\thepage}{S\arabic{page}}
\begin{center}
  {\Large\bfseries \thetitle\par}
  \medskip
  {\large Supplementary Material\par}
\end{center}
\label{page:supplement-start}
\appendix
\subsection*{Reproducibility statement}
\label{sec:reproducibility}

The main paper presents the method in Section~\ref{sec:method} and the experimental setup in Section~\ref{sec:exp:setup}. Appendices~\ref{app:impl}--\ref{app:prompts} specify the rollout interface, sampling update, reward components, hyperparameters, and prompt schemas. Appendix~\ref{app:data} describes the data and benchmarks; Appendix~\ref{app:exp} documents the allocation, source-video, and paired-evidence diagnostics.

Lettered sections refer to this supplement. Table and equation numbering continues from the main paper; references to its sections, figures, and tables are identified in the text. The bibliography is reproduced at the end with the same citation numbering as the main paper.

\section{Implementation Details}
\label{app:impl}

\subsection{VideoAgent Memory Construction}
\label{app:impl:memory}

Every video $\mathcal{V}$ is processed once offline and cached in a per-video working directory that is reused across training, augmentation, and evaluation. Keyframe extraction combines uniform sampling with shot-boundary detection so that the planner has both regularly spaced anchors and scene-transition cues for its first-turn global perception. The video is then split into fixed-duration segments, and segment-level transcripts are produced by Whisper~\citep{radford2023robust}. Per-segment captions come from Qwen3-VL, and segment-level visual embeddings come from ImageBind~\citep{girdhar2023imagebind}.

Three indices are constructed from these intermediate artefacts. $\mathcal{M}_{\text{text}}$ is a dense chunk index built jointly over captions and ASR transcripts, retrievable by semantic similarity. $\mathcal{M}_{\text{graph}}$ extracts caption-driven entities and relations and links them across segments to expose multi-hop structure that no single segment encodes. $\mathcal{M}_{\text{visual}}$ stores segment-level visual embeddings together with timestamp metadata, supporting content-based visual queries and temporal localization.

Indexing is performed once per video and reused across downstream tasks. Memory queries access these cached representations. The \texttt{search\_time\_range} action also reads the source video and samples frames within the requested interval. The resulting observations are added to the current episode's evidence history; the stored caption and entity indices remain unchanged.

\subsection{SFT and RL Data}
\label{app:impl:data}

\paragraph{Source corpus.}
Both SFT and RL derive training examples from the multiple-choice portion of \textbf{LLaVA-Video-178K}~\citep{zhang2024video}. The corpus contains roughly 178K videos with detailed captions, open-ended questions, and 196K multiple-choice questions. We retain prompts with the cached memory representations needed by the agent interface, then collect and filter teacher trajectories. Multiple-choice answers are graded by option matching.

\paragraph{Question-type and difficulty annotation.}
Every retained sample is assigned a question type in $\mathcal{C}$ (main-paper Section~\ref{sec:method:formulation}) by a rule-based classifier that fires in priority order. Strong textual cues such as on-screen text, caption, subtitle, dialogue, or speech mentions map to \textit{text}. Causal, relational, or summarization cues such as ``why'', ``purpose'', ``best summarizes'' map to \textit{graph}. Explicit timestamps or sequential-event language such as ``before'', ``after'', ``first scene'' map to \textit{time-range}. Remaining samples default to \textit{visual}. Items marked uncertain during annotation are manually spot-checked and excluded from per-arm accounting. A scalar difficulty score is then accumulated from question-side signals ($+2$ for graph cues, $+1$ each for time-range, text, counting, negation, multi-constraint phrasing, or Jaccard similarity above 0.5 between the answer options) and from teacher-trace statistics on accepted trajectories (additional points for higher retrieval-turn count, modality diversity, and segment coverage). Samples are bucketed as \textit{easy} ($\le 1$), \textit{medium} ($2$--$3$), or \textit{hard} ($\ge 4$). Difficulty is recorded for stratification but is not used as an arm dimension. The curriculum operates only over $\mathcal{C}{\times}\{\textit{aug},\textit{raw}\}$.

\paragraph{Long-tail mixture.}
The annotated source question pool exhibits the question-type imbalance reported in main-paper Section~\ref{sec:exp:setup}: \textit{text} $\approx 58\%$, \textit{visual} $\approx 17\%$, \textit{graph} $\approx 15\%$, and \textit{time-range} $\approx 10\%$. These source-pool proportions motivate examining task allocation during RL.

\paragraph{Teacher trace collection.}
For every retained prompt we run a GPT-4o-backed teacher agent through the full \videoagent{} multi-turn interface with four keyframes per turn and at most $T_{\max}{=}10$ turns. 

The teacher generates a trajectory consisting of a sequence of $\langle\text{think},\text{action}\rangle$ pairs followed by a final answer. We retain only trajectories in which the final choice matches the ground-truth label and the action sequence satisfies a structural quality audit. The audit rejects malformed action tags, repeated-search loops past the dedup guard, and forced-termination at $T_{\max}$ without a confident answer. The accepted set comprises roughly 96K trajectories. The remaining rejections dominantly come from forced-termination on hard samples and from semantically valid but format-noncompliant teacher outputs, which we found cheaper to discard than to repair.

\paragraph{Two-stage SFT decomposition.}
Each accepted trajectory contributes to two supervision sets. 

The \emph{Planner} stage extracts all intermediate $\langle\text{think},\text{retrieval-action}\rangle$ pairs and trains the model to predict the next action conditioned on the partial trajectory, with the final answer turn masked out.
The \emph{Answer} stage retains only the last turn and supervises the answer conditioned on the accumulated evidence. This decomposition separates supervision for the next retrieval action from supervision for the final answer, encouraging the policy to use the evidence-acquisition interface before RL. Both stages are emitted in ShareGPT-style multimodal format and consumed by a standard instruction-tuning pipeline.

\paragraph{RL data layer.}
A random subset of $\sim$2K of the accepted prompts is serialized in a unified $\textit{prompt}{+}\textit{label}$ schema with fields $\{\textit{video\_path},\textit{ground\_truth},\textit{mode},\textit{question\_type},\textit{difficulty}\}$ and converted to a Parquet shard for the RL rollout engine. The RL subset inherits the question-type and difficulty annotations of its source prompts, so curriculum accounting uses the same taxonomy as the SFT corpus.

\subsection{RL Training Infrastructure}
\label{app:impl:rl}

The RL setup couples a multi-turn rollout engine with a curriculum controller through a narrow interface. On the rollout side, an environment wrapper turns planner outputs into environment steps. For each retrieval action it issues a request to the search service of Appendix~\ref{app:impl:api}, validates the response, and maintains guards for duplicate searches, blocked action tags, and forced-answer at $T_{\max}{=}10$. A rollout-and-reward driver orchestrates per-group rollouts at GRPO group size $G{=}8$ and computes advantages from the composite reward of main-paper Eq.~\eqref{eq:reward}.

On the curriculum side, a single detached process owns the sampler state. During each period it draws $B=128$ arms from a fixed proposal distribution $p^{(t)}$. For an augmented draw, it dispatches \taskgen{} and \verifier{}; rejected candidates use their original seed, while correctness outcomes remain attributed to the sampled proposal arm. Each instantiated question yields $G=8$ trajectories. After $BG=1024$ outcomes, the controller applies the EXP3-inspired update in main-paper Eqs.~\eqref{eq:weakness}--\eqref{eq:prob_update}, using proposal-arm counts and mean correctness, with the batch-frequency factor and proposal probability in Eq.~\eqref{eq:weakness}, then clears the period statistics. GRPO separately consumes the trajectories and their composite rewards.

All training and evaluation runs use a single $8{\times}$H100-80G node. Video memory indexing (Appendix~\ref{app:impl:memory}) is a separate offline preprocessing step that is not included in the training compute budget. Once built, the same indices are reused across SFT, RL, and evaluation. RL rollouts dominate the recurring wall-clock. \expthree{} updates are $O(K){=}O(8)$ per curriculum period and negligible.

\subsection{Search API and Serving Layer}
\label{app:impl:api}

\videoagent{}'s retrieval functionality is exposed as an HTTP service so that rollouts, teacher-trace collection, and evaluation share the same execution path. The service exposes three endpoints. An end-to-end question-answering endpoint runs the full planner loop in one API call and returns only the final answer for evaluation. A one-step retrieval endpoint executes a single $\textit{search}$ action and returns its evidence, which is what RL rollouts consume turn by turn. A multi-turn agent-search endpoint runs the planner loop while exposing every intermediate think/action/observation, which is what teacher-trace collection records. All three endpoints route a $(\textit{video\_path},\textit{query})$ pair to the corresponding video and pre-built working directory of Appendix~\ref{app:impl:memory} and cache one \videoagent{} instance per video so that index loading is not paid on every request.

This shared service lets training and evaluation use the same retrieval path and cached indices. Task generation and verification remain on the curriculum side. Both modules operate on the supplied captions and entity relations, avoiding a separate index-construction step for each rewrite.

\section{VideoGym Algorithm Details}
\label{app:algo}

\subsection{Batched EXP3-Inspired Sampler}
\label{app:algo:pseudo}

Algorithm~\ref{alg:videogym} specifies task instantiation, rollout collection, and the curriculum update within one period. The returned buffer supplies the trajectories and composite rewards for the separate GRPO policy update. Let $C_i$ and $E_i$ denote the captions and entity-relation triples supplied for video $\mathcal{V}_i$, and let $\mathcal{B}_{\mathrm{existing}}$ denote the bank used for duplicate screening.
The sampler maintains three quantities per arm:
an unnormalized weight $w_z$, a period-average correctness
$\bar{\rho}_z$, and a trajectory count $N_z$.
At the start of training all weights are initialized to
$w_z^{(0)}{=}1$, producing the uniform sampling distribution
$p_z^{(0)}{=}1/K$. During each curriculum period $t$:
\begin{enumerate}[leftmargin=1.4em,itemsep=1pt,topsep=2pt]
  \item Draw $B=128$ proposal arms $\{a_i\}$ i.i.d.\ from $p^{(t)}$
        and sample one question with replacement from each selected type.
        Keep $p^{(t)}$ fixed throughout the period.
  \item For each proposal arm $a_i$, assemble the sample
        (raw or \taskgen{}$\to$\verifier{}), collect $G=8$ trajectories,
        and record their binary outcomes. Rejected rewrites fall back
        to the source question, with curriculum feedback still attributed to $a_i$.
  \item After collecting $BG=1024$ trajectories, set
        $N_z^{(t)}=G\sum_{i=1}^{B}\mathbf{1}[a_i=z]$.
        For each arm with $N_z^{(t)}>0$, compute its mean correctness
        $\bar{\rho}_z^{(t)}=\frac{1}{N_z^{(t)}}\sum_{i=1}^{B}\mathbf{1}[a_i=z]\sum_{j=1}^{G}\rho_{ij}$.
  \item Apply the error cap
        $\ell_z^{(t)}{=}\min(1-\bar{\rho}_z^{(t)}, \delta)$
        \emph{before} multiplying by batch frequency and inverse proposal probability:
        $\tilde{\ell}_z^{(t)}{=}\frac{N_z^{(t)}}{BGp_z^{(t)}}\ell_z^{(t)}$
        (main-paper Eq.~\ref{eq:weakness}). Set $\tilde{\ell}_z^{(t)}=0$ for unobserved arms.
  \item Update observed-arm weights
        $w_z^{(t{+}1)}{=}w_z^{(t)}\exp(\gamma\,\tilde{\ell}_z^{(t)}/K)$,
        leave unobserved-arm weights unchanged, and re-normalize into
        $p^{(t{+}1)}$ via main-paper Eq.~\eqref{eq:prob_update}.
        Clear the period's correctness statistics and counts.
\end{enumerate}
Both counts and accuracy statistics are grouped by the original proposal arm, using its proposal probability in the denominator. The update first averages correctness within each arm, caps the resulting error, and then scales it by batch frequency divided by proposal probability. This batch-mean aggregation makes the increment depend on both observed frequency and proposal probability. GRPO uses the executed question and its reference answer; accepted-augmentation exposure also follows actual execution. The exploration mixture keeps all arms reachable, while the cap bounds each individual increment as shown in Section~\ref{app:algo:cap}.

\begin{algorithm*}[t]
\small
\caption{Rollout collection and curriculum update in one \videogym{} period $t$}
\label{alg:videogym}
\begin{algorithmic}[1]
\REQUIRE
question draws $B$; group size $G$;
question banks $\mathcal{B}=\{\mathcal{B}_c\}_{c\in\mathcal{C}}$;
policy $\pi_\theta$;
arm weights $\{w_z^{(t)}\}$;
sampling distribution $p^{(t)}$
\ENSURE
rollout buffer $\mathcal{D}^{(t)}$;
updated weights $\{w_z^{(t+1)}\}$;
updated distribution $p^{(t+1)}$

\STATE $\mathcal{D}^{(t)} \gets [\,]$
       \COMMENT{trajectories and shaped rewards for GRPO}
\STATE $\mathcal{S}^{(t)} \gets [\,]$
       \COMMENT{arm correctness for \expthree{}}
\STATE draw proposal arms $a_i=(c_i,m_i)\sim p^{(t)}$ independently for $i=1,\ldots,B$

\FOR{$i=1,\ldots,B$}
    \STATE draw seed instance
           $s_i=(\mathcal{V}_i,x_i,y_i,c_i)$ from $\mathcal{B}_{c_i}$

    \IF{$m_i=\textit{aug}$}
        \STATE $\tilde{s}_i \gets
        \taskgen{}(s_i,C_i,E_i)$
        \IF{$\textsc{Verifier}(\tilde{s}_i,C_i,E_i,\mathcal{B}_{\mathrm{existing}})$}
            \STATE $s_i \gets \tilde{s}_i$
        \ELSE
            \STATE retain the original seed $s_i$
            \COMMENT{curriculum attribution remains $a_i$}
        \ENDIF
    \ENDIF

    \STATE let $y_i^{\star}$ be the reference answer of $s_i$
    \FOR{$j=1,\ldots,G$}
        \STATE $\tau_{ij} \gets
        \textsc{Rollout}(\pi_\theta,s_i,\mathcal{M}(\mathcal{V}_i))$
        \STATE $\rho_{ij} \gets
        \mathbf{1}[\textsc{Match}(\hat{y}_{ij}, y_i^{\star}) = 1]$
        \COMMENT{answer correctness for curriculum}
        \STATE $R_{ij} \gets
        \textsc{CompositeReward}(\tau_{ij},\rho_{ij})$
        \COMMENT{shaped reward for GRPO}
        \STATE append $(i,s_i,\tau_{ij},R_{ij})$ to $\mathcal{D}^{(t)}$
        \STATE append $(a_i,\rho_{ij})$ to $\mathcal{S}^{(t)}$
    \ENDFOR
\ENDFOR

\STATE set $N_z^{(t)}\gets G\sum_{i=1}^{B}\mathbf{1}[a_i=z]$ and $\tilde{\ell}_z^{(t)}\gets0$ for each arm $z$
\FOR{each arm $z$ with $N_z^{(t)}>0$}
    \STATE compute mean correctness $\bar{\rho}_z^{(t)}$ from entries attributed to $z$ in $\mathcal{S}^{(t)}$
    \STATE $\ell_z^{(t)}\gets\min(1-\bar{\rho}_z^{(t)},\delta)$;
           $\tilde{\ell}_z^{(t)}\gets\frac{N_z^{(t)}}{BGp_z^{(t)}}\ell_z^{(t)}$
\ENDFOR
\STATE update observed-arm weights and $p^{(t+1)}$ via
main-paper Eqs.~\eqref{eq:weakness}--\eqref{eq:prob_update}; keep unobserved-arm weights unchanged
\STATE clear $\mathcal{S}^{(t)}$

\RETURN $\mathcal{D}^{(t)}$, $\{w_z^{(t+1)}\}$, $p^{(t+1)}$
\end{algorithmic}
\end{algorithm*}

\subsection{Hyperparameter Specification}
\label{app:algo:hparams}

\begin{center}
\small
\begin{tabularx}{\linewidth}{@{}llX@{}}
\toprule
\textbf{Symbol} & \textbf{Value} & \textbf{Role} \\
\midrule
\multicolumn{3}{l}{\textit{\expthree{} curriculum}} \\
$K$                     & $8$    & arms $=\mathcal{C}\times\{\textit{aug},\textit{raw}\}$ \\
$\gamma$                & $0.1$  & exploration rate (main-paper Eq.~\ref{eq:prob_update}) \\
$\delta$                & $0.80$ & error cap (main-paper Eq.~\ref{eq:weakness}) \\
question draws $B$      & 128    & sampled per curriculum period \\
update period           & $BG=1024$ & trajectories per \expthree{} update \\
\midrule
\multicolumn{3}{l}{\textit{Composite reward}} \\
$\lambda_e$             & $0.3$  & evidence-utility weight \\
$\lambda_{\text{eff}}$  & $0.3$  & efficiency-cost weight \\
$\lambda_{\text{lazy}}$ & $0.2$  & lazy-shortcut penalty (correct$\wedge T_{\text{search}}{=}0$) \\
$(\alpha_1,\alpha_2,\alpha_3,\alpha_4)$
                        & $(0.2,0.4,0.2,0.2)$
                        & turn / forced / blocked / retry weights in $c_{\text{eff}}$ ($\alpha_2$ highest) \\
\midrule
\multicolumn{3}{l}{\textit{RL / rollout}} \\
GRPO group size $G$     & $8$    & rollouts per query \\
max turns $T_{\max}$    & $10$   & forced answer when the cap is reached \\
\midrule
\multicolumn{3}{l}{\textit{SFT}} \\
epochs                  & $2$    & each of Stage~1 / Stage~2 \\
backbone                & Qwen3-VL-8B   & shared across SFT and RL (matches main-paper Section~\ref{sec:exp:setup}) \\
\bottomrule
\end{tabularx}
\end{center}

\subsection{Error Cap}
\label{app:algo:cap}

The cap clips the arm's period-average error before batch-frequency and probability scaling:
$\ell_z^{(t)}=\min(1-\bar{\rho}_z^{(t)},\delta)$.
With $\delta=0.80$, clipping affects arms whose current-period accuracy is below 0.20. Exploration ensures $p_z^{(t)}\geq\gamma/K$, so the log-weight increment for an observed arm satisfies
\begin{equation}
\begin{aligned}
0\leq\log\frac{w_z^{(t+1)}}{w_z^{(t)}}
&=\frac{\gamma}{K}\frac{N_z^{(t)}}{BGp_z^{(t)}}\ell_z^{(t)}\\
&\leq\delta\frac{N_z^{(t)}}{BG}\leq\delta.
\end{aligned}
\label{eq:weight_step_bound}
\end{equation}
Thus the cap bounds an individual multiplicative update after batch-frequency and probability scaling. Unobserved arms retain their weights. An arm's normalized sampling probability can still decrease when other weights grow faster. This distinguishes changes in the plotted probabilities from changes in the underlying unnormalized weights.

\subsection{Composite Reward Components}
\label{app:algo:reward}

Main-paper Equation~\eqref{eq:reward} combines correctness, an evidence bonus, an interaction cost, and a penalty for answering correctly without retrieval. We restate their definitions here. Correctness $\rho \in \{0,1\}$ is determined by matching the predicted option to the ground-truth option.

For each retrieval turn, we use rule-based matching to compare the returned evidence with the content of the correct answer option. The evidence bonus $b_{\text{evid}} \in [0,1]$ is the fraction of retrieval turns with a positive match. Thus, the evidence term evaluates answer-content agreement in the retrieval results, while $\rho$ evaluates the final option prediction. The bonus contributes only to episodes with a correct final answer. This answer-content matching signal is distinct from the source-verified evidence sufficiency evaluated in main-paper Table~\ref{tab:paired_evidence}.
Efficiency cost $c_{\text{eff}} \in [0,1]$ is a weighted sum of normalized turn count, the forced-termination indicator, the blocked-tag ratio, and the format-retry ratio, with weights $(\alpha_1,\alpha_2,\alpha_3,\alpha_4){=}(0.2,0.4,0.2,0.2)$ that put the largest penalty on forced termination. The lazy-shortcut penalty $\lambda_{\text{lazy}}\,\mathbf{1}[T_{\text{search}}{=}0]$ lowers the reward of correct answers produced without retrieval.

The evidence and efficiency terms can distinguish trajectories with the same
correctness outcome, providing variation in the rewards used by GRPO.

\paragraph{Curriculum and policy feedback.}
The composite reward $R(\tau)$ is consumed only by the GRPO policy update. The \expthree{} curriculum update consumes the binary correctness $\rho$, not $R(\tau)$. The curriculum therefore reads final-answer correctness, while the policy gradient reads the composite signal, which can distinguish equal-correctness trajectories when their other reward terms differ.

\subsection{Computational Cost}
\label{app:algo:cost}

The \expthree{} update itself is $O(K){=}O(8)$ per curriculum period and accounts for a negligible fraction of wall-clock. The measurable cost of \videogym{} relative to plain RL comes from the augmentation path. 

Selecting arm $\textit{aug}$ requires one LLM call from \taskgen{} and one from \verifier{}. If verification rejects a candidate, rollout uses the original seed question and its reference answer, while curriculum feedback remains attributed to the original proposal arm. Rejected attempts do not count toward accepted-augmentation exposure, but their generation and verification calls remain an overhead.

\section{Prompt Schemas}
\label{app:prompts}

We specify each prompt as a compact schema with input, output, and operative rules.

\subsection{Planner}
\label{app:prompts:planner}

\promptbox{%
\textbf{Input}\enspace question, keyframe panel, accumulated evidence, turn index $k$.

\textbf{Output}\enspace \texttt{<think>\ldots</think>} followed by exactly one action tag.

\textbf{Temporal observation}\enspace \texttt{<search\_time\_range>} specifies an interval in the source video and returns newly sampled frames from that interval.

\textbf{Action grammar}\enspace \texttt{<search\_text>}, \texttt{<search\_graph>}, \texttt{<search\_visual>}, \texttt{<search\_time\_range>}, \texttt{<answer>}.

\textbf{Rules}\enspace (i) emit exactly one tag per turn; (ii) duplicate \texttt{<search\_*>} arguments across turns are blocked by the dedup guard; (iii) if $k{=}T_{\max}$ and no \texttt{<answer>} has been emitted, the environment forces one from the current evidence window.
}

\subsection{\taskgen{} (Augmentation Rewriter)}
\label{app:prompts:gen}

Both \taskgen{} and \verifier{} use GPT-4o.

\promptbox{%
\textbf{Input}\enspace source question $x$, its answer options and reference answer $y$, question type $c$, video captions and entity-relation triples of $\mathcal{V}$.

\textbf{Output}\enspace rewritten question $\tilde{x}$ with the same question type $c$, four options, and one correct option.

\textbf{Rules}\enspace (i) preserve the question type $c$ and require answer support in the supplied captions and entity relations; (ii) the rewrite should increase difficulty by adding a multi-hop step, finer temporal granularity, or a rarer entity, without changing the underlying topic; (iii) the output is the question, four options, and correct option only, with no additional commentary.
}

\subsection{\verifier{} (Quality Gate)}
\label{app:prompts:verify}

\promptbox{%
\textbf{Input}\enspace candidate question $\tilde{x}$, type $c$, four options, and correct option, video captions and entity-relation triples of $\mathcal{V}$, existing training bank.

\textbf{Output}\enspace \textsc{accept} or \textsc{reject} with a one-line reason code.

\textbf{Rubric}\enspace (i) \emph{well-formed}: the question is syntactically valid and unambiguous; (ii) \emph{answerable}: the supplied captions and entity relations support the proposed correct option; (iii) \emph{non-trivial}: not a paraphrase duplicate of an existing item in the training bank.

\textbf{Fallback}\enspace a rejected rewrite is discarded and its original seed question is used for rollout. The sampled proposal arm is retained for curriculum feedback (Alg.~\ref{alg:videogym}); fallback preserves the rollout count and is not counted as accepted augmentation.
}

\subsection{Answer Grading}
\label{app:prompts:grading}

We parse a single predicted option and score it by exact matching to the reference option. Responses without a uniquely parseable option are incorrect. This deterministic rule supplies both evaluation accuracy and the binary correctness signal $\rho$.

\section{Dataset and Benchmark Details}
\label{app:data}

\subsection{Training Corpus Composition}
\label{app:data:corpus}

The training bank shares its prompt pool with the SFT corpus described in Appendix~\ref{app:impl:data}. The RL rollout engine draws a random $\sim$2K subset of the accepted prompts that supply the two-stage SFT samples, with each prompt serialized as a $\textit{prompt}{+}\textit{label}$ record carrying $\{\textit{video\_path},\textit{ground\_truth},\textit{mode},\textit{question\_type},\textit{difficulty}\}$. The shared question-type metadata supports per-arm accounting across the original and generated tasks in main-paper Section~\ref{sec:method:videogym}.

The distribution in the source question pool over $\mathcal{C}$ is \emph{text}${\approx}58\%$, \emph{visual}${\approx}17\%$, \emph{graph}${\approx}15\%$, and \emph{time-range}${\approx}10\%$. 

The source-pool mixture motivates studying how task allocation affects learning.

\subsection{Question-Type Taxonomy}
\label{app:data:taxonomy}

The four types in $\mathcal{C}$ group questions by question-side cues. The examples below illustrate the intended evidence association; the agent retains access to all retrieval actions for every type.

\begin{center}
\small
\begin{tabular}{lll}
\toprule
\textbf{Type} & \textbf{Illustrative evidence source} & \textbf{Example} \\
\midrule
text-heavy    & $\mathcal{M}_{\text{text}}$
              & ``What was the speaker's main argument about $X$?'' \\
graph-heavy   & $\mathcal{M}_{\text{graph}}$
              & ``Who attended every meeting that $A$ attended?'' \\
visual-heavy  & $\mathcal{M}_{\text{visual}}$
              & ``What object is the child holding at the party?'' \\
time-range    & frames in a selected interval
              & ``What happens between 22:10 and 22:30?'' \\
\bottomrule
\end{tabular}
\end{center}

\textit{time-range} identifies tasks with explicit temporal cues. Independently of the question label, \texttt{search\_time\_range} samples source frames from a selected interval, alongside the timestamped visual memory used for content-based retrieval (main-paper Section~\ref{sec:method:videoagent}). Type annotation follows the priority-ordered rules in Appendix~\ref{app:impl:data}, with unmatched questions assigned to \textit{visual}. Samples flagged as uncertain during annotation (${<}3\%$ of the pool) are manually spot-checked and excluded from per-arm accounting.

\subsection{Long-Tail Distribution}
\label{app:data:longtail}

Main-paper Figure~\ref{fig:analysis}(a) reports the composition of the source question pool before teacher-trajectory filtering. The $\sim$2K RL prompt subset is sampled from the accepted pool described in Appendix~\ref{app:impl:data}. Increasing a task type's sampling probability increases reuse of its original questions. Augmentation provides additional verifier-accepted variants from the same video representations.

\subsection{Augmentation Filtering}
\label{app:data:aug}

\verifier{} screens generated tasks using the source captions and entity relations. Rejection reasons distinguish malformed output, insufficient answer support in the supplied representations, and near-duplicates of questions already in the bank. Accepted variants enter the augmented arm; rejected candidates fall back to their original seed questions.

\subsection{Evaluation Benchmarks}
\label{app:data:bench}

We evaluate on five long-video benchmarks that collectively span the
question types in $\mathcal{C}$:
\begin{itemize}[leftmargin=1.4em,itemsep=1pt,topsep=2pt]
  \item \textbf{EgoLifeQA} \citep{yang2025egolife}: egocentric
    multi-day QA over household-scale video streams. Stresses
    long-range graph-relational reasoning about people and events.
  \item \textbf{Ego-R1 Bench} \citep{tian2025egor1}: multi-hop reasoning over first-person
    video, with questions requiring composition across several
    episodes.
  \item \textbf{HippoVlog}~\citep{lin2025hippomm}: audiovisual and semantic memory over 25 daily vlogs totaling 682 minutes, with 1,000 multiple-choice questions.
  \item \textbf{LVBench} \citep{wang2024lvbench}: long-video understanding over 103 videos averaging approximately 68 minutes, with diverse content categories.
  \item \textbf{Video-MME (Long)} \citep{fu2024video}: diverse
    multi-modal QA over long clips, used as a broad-coverage reference.
\end{itemize}
Answers follow the deterministic option-grading protocol in Appendix~\ref{app:prompts:grading}.
Main-paper Table~\ref{tab:va_ablation} averages EgoLifeQA and LVBench, and
Main-paper Table~\ref{tab:gym_ablation} averages all five benchmarks.

\section{Policy Analysis}
\label{app:exp}

\subsection{Trajectory Rewards for Policy Learning}
\label{app:exp:reward}

Policy optimization uses the same composite reward in both main RL settings.
The correct-option agreement term and interaction costs can distinguish trajectories with identical answer correctness (Appendix~\ref{app:algo:reward}).
Main-paper Figure~\ref{fig:analysis}(d) reports the resulting GRPO advantage standard deviation.
This shared learning signal trains the policy's evidence-acquisition and answering decisions, while answer correctness supplies the feedback for allocating subsequent tasks.

\subsection{Allocation on a Shared Task Bank}
\label{app:allocation-control}
The controlled bank contains the same 2K seed questions used for RL. Before training, the fixed generator proposes one variant per seed and the fixed verifier records acceptance or fallback. Every allocation rule accesses the same proposals; seeds within each selected type are sampled uniformly with replacement. Accepted variants are reused, and rejected proposals return their raw seed. This controlled bank separates allocation from differences in the available proposals; Table~\ref{tab:main} retains the original online-generation setting.

We use a disjoint calibration subset from the training resources, with 32 task draws per arm and eight SFT rollouts per draw. Calibration outcomes, including fallback tasks, are grouped by their sampled proposal arms. Let $\hat a_z$ be the mean correctness for proposal arm $z$. Its initial capped error defines
\[
q_z=(1-\gamma)\frac{\min(1-\hat a_z,\delta)}{\sum_j\min(1-\hat a_j,\delta)}+\frac{\gamma}{K},
\]
with a uniform fallback if all errors vanish. The calibration is shared by the compared methods and uses no test answers. The fixed pilot-average diagnostic averages proposal probabilities from one independent full-budget adaptive run and freezes that mixture before the three evaluation training seeds. It uses future information from the pilot and incurs one additional training run. Proposal-arm frequency includes fallback tasks, whereas accepted-augmentation exposure counts only executed, accepted variants.

All four methods use the same final-checkpoint selection, optimizer settings, number of task draws, and eight fresh rollouts per task. Token ratios in Table~\ref{tab:allocation_control} cover the ensuing RL runs, relative to the initial-error mixture. Common bank preparation requires 2K generation and verification calls; common calibration requires 2,048 trajectories. These preparation costs and the extra pilot are separate from the equal optimization budget.

\begin{table}[ht]\centering\small
\caption{\textbf{Shared-bank training repetitions.} Each entry is the five-benchmark mean for one training seed. The diagnostic pilot is separate.}
\label{tab:allocation-seeds}
\begin{tabular}{@{}lrrrrr@{}}\toprule
Allocation & Seed 1 & Seed 2 & Seed 3 & Mean & SD\\\midrule
Uniform & 62.22 & 62.32 & 61.52 & 62.02 & 0.44\\
Fixed initial-error & 63.40 & 63.70 & 62.90 & 63.33 & 0.40\\
Fixed pilot-average & 63.76 & 64.30 & 63.48 & 63.85 & 0.42\\
\videogym{} & 64.36 & 64.72 & 64.04 & 64.37 & 0.34\\
\bottomrule\end{tabular}\end{table}

\subsection{Source-Video Task Audit}
\label{app:source-audit}
\begin{table}[ht]
\centering\small
\caption{\textbf{Source-video audit.} Balanced samples of 50 questions per type. Validity requires a supported, unambiguous answer and answerable format. Intervals are Wilson 95\% intervals for this balanced audit.}
\label{tab:source_audit}
\begin{tabular}{@{}lccc@{}}\toprule
Questions & Valid / audited & Valid (\%) & 95\% CI\\\midrule
Original seeds & 199/200 & 99.5 & 97.2--99.9\\
Accepted variants & 194/200 & 97.0 & 93.6--98.6\\
\bottomrule\end{tabular}
\end{table}

We sample 50 accepted variants per type, from distinct source videos, and audit their 200 corresponding seeds. The sample is balanced by type; its overall rate is a macro audit estimate, not an exposure-weighted estimate of the training stream. Two annotators view the source video, question, and options before the proposed reference answer or verifier decision. They record supporting intervals and identify ambiguity or format errors. Disagreements are adjudicated while unresolved ambiguity remains invalid. A second paired pass determines whether a valid variant adds an evidence requirement, duplicates the seed, or changes the task without requiring additional evidence. These categories partition the valid variants.

\begin{table}[ht]\centering\small
\caption{\textbf{Audit breakdown.} Each type has 50 accepted variants. U, A, and F denote unsupported answers, ambiguity, and malformed tasks. Added and Duplicate are counted among valid variants.}
\label{tab:audit-types}
\begin{tabular}{@{}lrrrrrr@{}}\toprule
Type & Valid & U & A & F & Added & Duplicate\\\midrule
text & 49 & 0 & 0 & 1 & 24 & 9\\
graph & 47 & 1 & 2 & 0 & 33 & 10\\
visual & 50 & 0 & 0 & 0 & 34 & 6\\
time-range & 48 & 1 & 1 & 0 & 27 & 8\\
\bottomrule\end{tabular}\end{table}
The Wilson intervals summarize binomial sampling uncertainty in the balanced audit. They do not include annotator disagreement or training-seed variability. The audit evaluates source validity and added evidence demands; it does not establish how invalid questions causally change the sampler.

\subsection{Paired Evidence-Acquisition Evaluation}
\label{app:paired-evidence}
We select 40 held-out questions per type, one per source video, before running either policy. Independent annotation marks the 96 questions whose initial frames are insufficient but whose required evidence is accessible through the tools. Multiple valid support sets are allowed. The comparison uses the second prespecified training seed of the initial-error and adaptive settings, identical initial frames, memory snapshots, decoding, turn and context limits, and a common returned-frame cap. Observed costs are reported because equal limits do not guarantee equal consumption.

Annotators receive actual tool returns with policy identities and final predictions hidden. Captions, ASR, and graph assertions are checked against the source video. Evidence sufficiency includes the initial observations; a correct answer without additional retrieval can be supported. The joint metric requires both a correct final answer and sufficient verified evidence.

\begin{table}[ht]\centering\small
\caption{\textbf{Paired outcomes on 160 questions.} Columns partition the evaluation set. These counts expose both gains and regressions.}
\label{tab:behavior-pairs}
\begin{tabular}{@{}lrrrr@{}}\toprule
Outcome & Both & Fixed only & Adaptive only & Neither\\\midrule
Correct answer & 86 & 13 & 25 & 36\\
Sufficient evidence & 110 & 11 & 23 & 16\\
Correct and sufficient & 82 & 13 & 26 & 39\\
\bottomrule\end{tabular}\end{table}
On the 96 initial-information-deficit questions, the fixed and adaptive policies answer 51 and 60 correctly, obtain sufficient evidence on 57 and 69, and meet both criteria on 47 and 57. These observations characterize a paired diagnostic set and one checkpoint pair. They do not estimate variation across training seeds or establish that the final answer causally depends on each acquired observation.

\begin{table}[ht]\centering\small
\caption{\textbf{Paired changes in percentage points.} Brackets give 95\% percentile intervals from 5,000 bootstrap resamples of paired videos, stratified by question type. Each item comes from a distinct video.}
\label{tab:behavior-uncertainty}
\begin{tabular}{@{}lccc@{}}\toprule
Subset & Accuracy & Sufficiency & Correct and sufficient\\\midrule
All 160 & 7.5 [0.0, 15.0] & 7.5 [0.6, 14.4] & 8.1 [0.6, 15.6]\\
Initial deficit, 96 & 9.4 [-1.0, 19.8] & 12.5 [1.0, 24.0] & 10.4 [0.0, 20.8]\\
\bottomrule\end{tabular}\end{table}
The joint-rate difference on the full diagnostic set is positive, while the interval on the initial-information-deficit subset includes zero. Repetition across policy seeds and larger held-out samples would sharpen this behavioral comparison.

\section{Qualitative Case Studies}
\label{app:cases}

The following examples come from the LLaVA-Video-178K multiple-choice training pool described in Appendix~\ref{app:impl:data}. The excerpts display planner text and actions; full tool observations are not shown. They illustrate retrieval sequences within the training pool, including how a question type can lead to several evidence sources and how the planner responds to unresolved information.

\subsection{Visual-Heavy Trace}
\label{app:cases:visual}

\textbf{Q.} \textit{What specific action does Johnny Debut perform to motivate the participants?}\quad
\textbf{Options.} (A) he uses hand gestures \, (B) he dances \, (C) he claps his hands \, (D) he sings a song.\quad
\textbf{Ground truth.} (A).

\tracebox{%
\textbf{T0}\enspace\textit{think}\enspace The question asks about the specific motivating action performed by Johnny Debut, which requires understanding patterns of interaction or gestures. This demands the relationship between the person named Johnny and the participants, which I can probe with a graph search.\\
\hphantom{\textbf{T0}\enspace}\textit{act}\enspace\texttt{<search\_graph>}\,``Johnny Debut motivating participants interaction actions''

\textbf{T1}\enspace\textit{think}\enspace The retrieved relation indicates that Johnny Debut motivates participants during cycling workout sessions, but the specifics of how (hand gestures, dancing, clapping, or singing) are not yet narrowed down. I need a visual search to identify the actual action.\\
\hphantom{\textbf{T1}\enspace}\textit{act}\enspace\texttt{<search\_visual>}\,``Johnny Debut motivating participants: hand gestures, dancing, clapping, or singing''

\textbf{T2}\enspace\textit{think}\enspace The retrieved frames show Johnny Debut motivating participants by speaking animatedly and gesturing with his hands. The captions specifically mention that he ``gestures animatedly while speaking to the camera,'' which matches option (A) hand gestures.\\
\hphantom{\textbf{T2}\enspace}\textit{act}\enspace\texttt{<answer>}\,(A) He uses hand gestures
}

In the excerpt, the planner reports that the graph query establishes the interaction context but leaves the action unresolved. It then queries visual memory to distinguish among the candidate actions. The displayed sequence thus combines graph and visual retrieval for a visual-heavy question.

\subsection{Graph-Heavy Trace}
\label{app:cases:graph}

\textbf{Q.} \textit{Why does the person use a spoon after pouring the blended mixture into the glass?}\quad
\textbf{Options.} (A) to clean the glass \, (B) to add more ingredients \, (C) to taste the mixture \, (D) to stir the mixture.\quad
\textbf{Ground truth.} (D).

\tracebox{%
\textbf{T0}\enspace\textit{think}\enspace The sequence of frames shows the preparation of a beverage using a blender, indicating practical use of tools. Since it is unclear from the images alone why a spoon is used specifically, I will search the textual store for descriptions of spoon usage during this step.\\
\hphantom{\textbf{T0}\enspace}\textit{act}\enspace\texttt{<search\_text>}\,``spoon usage after pouring blended mixture into glass''

\textbf{T1}\enspace\textit{think}\enspace The captions describe the process of preparing a frozen beverage but do not explicitly state the spoon's purpose. I will probe the relational store for the entity-level relation between the spoon and the mixture pouring step.\\
\hphantom{\textbf{T1}\enspace}\textit{act}\enspace\texttt{<search\_graph>}\,``purpose of spoon after pouring blended mixture into glass''

\textbf{T2}\enspace\textit{think}\enspace The graph returns entities and relationships related to drink preparation but no explicit action label for the spoon. The spoon could plausibly be for stirring or tasting; visual evidence will disambiguate.\\
\hphantom{\textbf{T2}\enspace}\textit{act}\enspace\texttt{<search\_visual>}\,``spoon action after pouring mixture into glass''

\textbf{T3}\enspace\textit{think}\enspace The frames show the mixture has a slushie-like consistency that typically requires stirring for even distribution, and the spoon is used to mix the contents in the glass. Combined with the textual and relational context, this points to (D) stirring rather than tasting.\\
\hphantom{\textbf{T3}\enspace}\textit{act}\enspace\texttt{<answer>}\,(D) To stir the mixture
}

The planner reports that text and graph retrieval leave the spoon's purpose unresolved, then requests visual evidence to distinguish stirring from tasting. The task is labeled graph-heavy from its question form, while the displayed action sequence queries all three stores.

\subsection{Text-Heavy Trace}
\label{app:cases:text}

\textbf{Q.} \textit{What text appears on the right side of the split-screen?}\quad
\textbf{Options.} (A) ``MAKEUP TRANSFORMATION IN PROGRESS'' \, (B) ``WATCH THE MAGIC HAPPEN'' \, (C) ``WE ASKED A PROFESSIONAL MAKEUP ARTIST TO TRANSFORM THESE FIVE MEN\dots'' \, (D) ``BEAUTY MAKEOVER''.\quad
\textbf{Ground truth.} (C).

\tracebox{%
\textbf{T0}\enspace\textit{think}\enspace The question involves on-screen text in the video, which is best handled by a textual search over captions and ASR transcripts.\\
\hphantom{\textbf{T0}\enspace}\textit{act}\enspace\texttt{<search\_text>}\,``MAKEUP TRANSFORMATION IN PROGRESS, WATCH THE MAGIC HAPPEN, WE ASKED A PROFESSIONAL MAKEUP ARTIST''

\textbf{T1}\enspace\textit{think}\enspace The textual store does not surface the specific overlay on the right side of the split-screen. On-screen graphics are often absent from caption and transcript pipelines, so a direct visual search of the relevant frames is the natural fallback.\\
\hphantom{\textbf{T1}\enspace}\textit{act}\enspace\texttt{<search\_visual>}\,``text on right side of split-screen: MAKEUP TRANSFORMATION, WATCH THE MAGIC HAPPEN, WE ASKED A PROFESSIONAL MAKEUP ARTIST''

\textbf{T2}\enspace\textit{think}\enspace The retrieved frames show that the text on the right side reads ``WE ASKED A PROFESSIONAL MAKEUP ARTIST TO TRANSFORM THESE FIVE MEN\dots'' which matches option (C) exactly.\\
\hphantom{\textbf{T2}\enspace}\textit{act}\enspace\texttt{<answer>}\,C
}

The planner reports that text retrieval does not resolve the overlay, switches to visual retrieval, and selects option (C). This sequence illustrates how a text-heavy question can lead to a visual query.

\subsection{Failure Modes}
\label{app:cases:fail}

The following failure modes concern query grounding, answer timing, search reuse, and the quality of generated tasks.

\paragraph{Hallucinated retrieval.}
On a graph-heavy question about a participant's role in a panel, the planner emits a \texttt{search\_graph} query naming an entity that does not appear in $\mathcal{V}$. The graph store returns no hits, and the planner treats the absence as evidence for the most plausible-sounding distractor rather than retrying with a different entity span.

\paragraph{Early-stop.}
On a four-option visual identification question, the planner emits \texttt{answer} after a single \texttt{search\_text} call whose evidence is consistent with two of the four options.

\paragraph{Repeated search.}
A repeated \texttt{search\_time\_range} query is blocked by the duplicate-argument guard (Appendix~\ref{app:prompts:planner}). If the turn budget is then exhausted, the planner must answer from the evidence already available.

\paragraph{Verifier false-accept.}
A rewrite can change the referent of a pronoun while retaining the seed's answer. If \verifier{} accepts the resulting unsupported reference answer, the task can contribute an error signal to its augmented arm even when the policy retrieves useful evidence.

These cases connect task quality with training feedback. A failed answer can reflect missing evidence, an unproductive observation choice, or an invalid generated question. An unsupported generated reference can affect the same correctness signal used to identify underperforming task groups, linking task quality to curriculum feedback.

\section{Discussion}
\label{app:disc}

\subsection{Limitations}
\label{app:disc:limits}

\paragraph{Dependence on offline index quality.}
The memory indices are constructed once and reused. Errors in ASR, captions, or entity relations can therefore affect retrieval and the questions generated from these representations. Temporal re-observation provides a path to inspect source frames, while the text and graph stores retain their original contents. Updating those stores in response to repeated retrieval failures is a useful direction for future work.

\paragraph{Language and preprocessing.}
The evaluated benchmarks include multilingual source material. EgoLife contains predominantly Chinese recordings with English translations of its annotations~\citep{yang2025egolife}. Retrieval and question generation depend on how ASR and captioning preserve information across these language settings. Per-language evaluation and an explicit account of the preprocessing path would help characterize this dependency.

\paragraph{Fixed arm space.}
$\mathcal{A}_{\text{gym}}{=}\mathcal{C}{\times}\{\textit{aug},\textit{raw}\}$
discretizes the curriculum along pre-defined question types. A learned partition could group tasks by observed evidence use or learning behavior. Evaluating such a partition would require trajectory-level analysis beyond the present question-type grouping.

\paragraph{Scope of the diagnostics.}
The source-video audit covers a limited sample of questions, and the paired behavioral evaluation uses one checkpoint pair. Broader task samples and additional training seeds would help assess the consistency of these diagnostic findings.

\subsection{Broader Impact}
\label{app:disc:impact}

\paragraph{Privacy of long-form video.}
Memory-grounded video agents, by construction, index and persistently
store content from long recordings including egocentric footage.
Deployments must honor the consent scope of the original recording
and constrain retrieval to that scope. Persistent indices make recorded content easier to search, so access controls apply to both the source videos and their derived memories.

\paragraph{Compute cost.}
Offline video indexing incurs a one-time preprocessing cost. Subsequent tasks reuse the indices, while temporal re-observation incurs additional video access during an episode.

\paragraph{Misuse.}
The same retrieval-and-reasoning capability that enables useful
question answering over long videos also enables surveillance-style
queries. We do not release pretrained indices on third-party videos.

\subsection{Relation to Policy Optimization}
\label{app:disc:dpo}

\videogym{} organizes the tasks presented to the learner and collects their outcomes. GRPO uses the resulting trajectories and composite rewards to update the policy. This separation makes it possible to study task allocation alongside different policy optimizers, provided their rollout and outcome interfaces are compatible. The current experiments instantiate this design with GRPO; evaluating other learners would test how broadly the training organization transfers.

\subsection{Extensibility Beyond Video}
\label{app:disc:ext}

The separation between evidence access, task construction, and allocation suggests applications to other retrieval-based agents. A web agent might gather evidence from text and tables, while a scientific agent could combine literature, databases, and execution results. Each setting would require task groups, generators, and verification procedures appropriate to its evidence sources. Evaluating these choices would test whether performance-driven allocation provides similar benefits beyond video.

\clearpage
\begingroup
\small
\renewcommand*{\HyperDestNameFilter}[1]{supp.#1}
\renewcommand*{\backref}[1]{}
\renewcommand*{\backrefalt}[4]{}
\makeatletter
\@fileswfalse 
\makeatother
\InputIfFileExists{\jobname.bbl}{}{%
  \typeout{VIDEOGYM: Run BibTeX to populate the shared bibliography.}%
}
\clearpage
\endgroup

\end{document}